\documentclass[11pt]{article}
\usepackage[utf8]{inputenc}
\usepackage{amsmath}
\usepackage{graphicx}
\usepackage{hyperref}
\usepackage[table,xcdraw]{xcolor}
\usepackage{pifont}

\title{\Large \textbf{16-811 Math Fundamentals for Robotics}\\
\Large Comparison of Optimization Methods in Optical Flow Estimation\\
\large Final Report, Fall 2015}
\author{\textbf{Noranart Vesdapunt}
\\ Master of Computer Vision
\\\small Carnegie Mellon University
\\\texttt nvesdapu@andrew.cmu.edu
\and
\textbf{Utkarsh Sinha}
\\Master of Computer Vision
\\\small Carnegie Mellon University
\\\texttt usinha@andrew.cmu.edu}
\date{December 3, 2015}

\usepackage[margin=1.3in]{geometry}

\begin{document}
\setcounter{page}{0}

\maketitle
\tableofcontents
\newpage

\section{Introduction}

Optical flow estimation is a widely known problem in computer vision introduced by \cite{1} to describe the visual perception of human by stimulus objects. Estimation of optical flow model can be achieved by solving for the motion vectors from region of interest in the the different timeline. Motion vectors are defined by the relative velocity between object and observer. The standard method of solving this problem is least square estimation using Singular Value Decomposition (SVD). However, optical flow has more unknown variables than known equation. Therefore, variation of solving this problem is about empirically defines constraints to reduce unknown variables. In this paper, we assumed slightly uniform change of velocity between two nearby frames, and solve the optical flow problem by traditional method, Lucas-Kanade \cite{2}. This method performs minimization of errors between template and target frame warped back onto the template. Solving minimization steps requires optimization methods which have diverse convergence rate and error. We explored first and second order optimization methods, and compare their results with Gauss-Newton method in Lucas-Kanade. Since direct observation of optical flow required distinguishing the movement of single pixel, we instead evaluated optical flow by its application, object tracking. We fixed the size of tracking box and computed the errors by Euclidean distance. We generated 105 videos with 10,500 frames by synthetic objects, and 10 videos with 1,000 frames from real world footage. Our experimental results could be used as tuning parameters for Lucas-Kanade method.
\newline

\section{Method}
\subsection{Optical Flow}
Optical flow is the model of object motion in a visual scene caused by the motion from a camera or an object within the scene. In Computer Vision, estimation of optical flow is achieving by calculation of motion between two image frames at time $t$ and $t+\Delta t$. Since we are interesting in digital image, we estimate optical flow at each pixel in 2D space. We setup the equation between pixel of two image frames as following:
\begin{equation}
I(x_2,y_2,t_2)=I(x_1+\Delta x,y_1+\Delta y,t_1+\Delta t)
\end{equation}
The Taylor series of Equation (1) can be given:
\begin{equation}
I(x+\Delta x,y+\Delta y,t+\Delta t)=I(x,y,t)+\frac{\partial I}{\partial x}\Delta x+\frac{\partial I}{\partial y}\Delta y+\frac{\partial I}{\partial t}\Delta t+...
\end{equation}
For simplicity, we considered only the first order term of Taylor series.
\begin{equation}
I(x+\Delta x,y+\Delta y,t+\Delta t)-I(x,y,t)=\frac{\partial I}{\partial x}\Delta x+\frac{\partial I}{\partial y}\Delta y+\frac{\partial I}{\partial t}\Delta t
\end{equation}
The left term is the difference of intensity between two frames. If we warped the second frame onto the first frame, the best optical flow estimation will be the minimization of error between these frames. Hence, we setup the difference as 0, and solve for least square error.
\begin{equation}
0=\frac{\partial I}{\partial x}\Delta x+\frac{\partial I}{\partial y}\Delta y+\frac{\partial I}{\partial t}\Delta t
\end{equation}
\begin{equation}
0=\frac{\partial I}{\partial x}\frac{\Delta x}{\Delta t}+\frac{\partial I}{\partial y}\frac{\Delta y}{\Delta t}+\frac{\partial I}{\partial t}\frac{\Delta t}{\Delta t}
\end{equation}
\begin{equation}
I_xV_x+I_yV_y=-I_t
\end{equation}
Equation (6) shows that we want to solve two unknown variables ($I,V$) with one equation. Therefore, state of the art for solving this problem is about setting up constraints and configurations to reduce degree of freedom. In our paper, we chose to further explore the traditional method, Lucas Kanade.
\newline

\subsection{Lucas Kanade}
Lucas Kanade method assumes the small and constant velocity between two nearby frames. We want to minimize the sum of squared error between template $T$ (or the first frame) and the second frame $I$ warped back onto the template. We setup the cost function as following:
\begin{equation}
\sum\limits_{x,y} [I(W(x,y;p))-T(x,y)]^2
\end{equation}
Minimization of Equation (7) is a non-linear optimization because pixel values $I(x,y)$ are independent to coordinates $x,y$. Therefore, this method is optimized by solving descent step ($\Delta p$).
\begin{equation}
\sum\limits_{x,y} [I(W((x,y;p)+\Delta p))-T(x,y)]^2
\end{equation}
And we update the step with
\begin{equation}
p_{t+\Delta t}=p_t+\Delta p
\end{equation}
The first order term of Taylor series for Equation (8) is
\begin{equation}
\sum\limits_{x,y} [I(W(x,y;p)+\nabla I\frac{\partial W}{\partial p}\Delta p-T(x,y)]^2
\end{equation}
We assume the motion is pure translation.
\begin{equation}
W((x,y;p)+\Delta p))=
  \left[\begin{matrix}
    x+p_{1}\\
    y+p_{2}\\
  \end{matrix}\right]
\end{equation}
\begin{equation}
\frac{\partial W}{\partial p}=
  \left[\begin{matrix}
    1 & 0\\
    0 & 1\\
  \end{matrix}\right]
\end{equation}
\cite{2} demonstrate that Lucas Kanade solve $\Delta p$ by Gauss-Newton Method where $\Delta p$ is as following:
\begin{equation}
\Delta p=H^{-1}\sum\limits_{x,y}[\nabla I\frac{\partial W}{\partial p}]^T[T(x,y)-I(W(x,y;p))]
\end{equation}
From Equation (12)
\begin{equation}
\Delta p=H^{-1}\sum\limits_{x,y}[\nabla I]^T[T(x,y)-I(W(x,y;p))]
\end{equation}
where H is the approximation of Hessian matrix from Gauss-Newton Method
\begin{equation}
H=\sum\limits_{x,y}[\nabla I\frac{\partial W}{\partial p}]^T[\nabla I\frac{\partial W}{\partial p}]
\end{equation}
\begin{equation}
H=\sum\limits_{x,y}[\nabla I]^T[\nabla I]
\end{equation}
Equation (14) shows that we need to iteratively solve Hessian matrix for every change of frames which is computationally expensive. Hence, we looked at alternative ways of solving the optimization problems and compared the results.
\newline

\subsection{Gradient Descent}
Gradient descent solves the optimization problem by first order algorithm. Given the gradient descent equation \cite{3}:
\begin{equation}
b=a-\eta\nabla F(a)
\end{equation}
In our case, we want to solve $\Delta p$ in Equation (8). We adjusted Equation (17) as following:
\begin{equation}
p_{t+\Delta t}=p_t+\eta\sum\limits_{x,y}[\nabla I]^T [T(x,y)-I(W(x,y;p)]
\end{equation}
\begin{equation}
\Delta p=\eta\sum\limits_{x,y}[\nabla I]^T[T(x,y)-I(W(x,y;p)]
\end{equation}
Update rules from Equation (19) demonstrates the parallelism with Equation (14). In Gauss-Newton Method, we used approximation of  Hessian matrix as $\eta$. Therefore, comparison between these two optimization means dynamically compute $\eta$, or fix it as constant value. We ran an experiment on several constant $\eta$ in experimental section and compare results with traditional method.
\\

\subsection{Conjugate Gradient Descent}
Conjugate gradient descent solves the non-linear optimization problem by first order algorithm similar to gradient descent. However, instead of descenting with fix $\eta$ step, this method makes use of the concept of conjugate vectors. Two vectors have the property of conjugate vector, if and only if:
\begin{equation}
u^TAv=0
\end{equation}
From gradient descent method, Equation 19:
\begin{equation}
\nabla F(x,y)=[\nabla I]^T\sum\limits_{x,y}[T(x,y)-I(W(x,y;p)]
\end{equation}
Then we iterate the following steps \cite{4}:
\begin{equation}
s_n=-\nabla F(x,y)+ \beta _n s_{n-1}
\end{equation}
\begin{equation}
\alpha_n=min(F((x,y)+\alpha s_n))
\end{equation}
\begin{equation}
\Delta p=\alpha_n s_n
\end{equation}
Equation (24) defines the update rule as auto-calculated $\eta$ step by solving line search optimization. However, in our scenario, solving this problem in 2D dimension by brute force required highly computational cost. We avoided the bottle neck by setting up a constant step $\eta$ in the same fashion as gradient descent. We tested various constant steps in experimental section. $s_n$ in Equation (22) is similar to $\nabla F$ in gradient descent with an additional boost by $\beta$ from conjugate vector. Solving $\beta$ can be done in several ways, therefore, we chose the most popular method including:
\begin{itemize}
  \item Fletcher Reevee
\begin{equation}
\beta_n=\frac{\Delta p_n^T \Delta p_n}{\Delta p_{n-1}^T \Delta p_{n-1}}
\end{equation}
  \item Polak Ribire
\begin{equation}
\beta_n=\frac{\Delta p_n^T(\Delta p_n-\Delta p_{n-1})}{\Delta p_{n-1}^T \Delta p_{n-1}}
\end{equation}
  \item Hestenes-Stiefel
\begin{equation}
\beta_n=-\frac {\Delta p_n^T(\Delta p_n-\Delta p_{n-1})}{s_{n-1}^T(\Delta p_n-\Delta p_{n-1})}
\end{equation}
  \item Dai Yuan
\begin{equation}
\beta_n=-\frac{\Delta p_n^T \Delta p_n}{s_{n-1}^T(\Delta p_n-\Delta p_{n-1})}
\end{equation}
\end{itemize}   
~\newline
If the given function is quadratic, these formulas will be equivalent. However, in non-linear optimization, each $\beta$ depends on given function in the heuristics fashion. Solving for $\alpha,\beta, s_n$ required higher computational complexity than gradient descent. However, the boosting from these extra parameters will be likely to result in faster rate of convergence.

~\\
\subsection{Newton's Method}
Newton's method solves the optimization problem by second order algorithm. Given the Newton's method equation \cite{5}:
\begin{equation}
b=a-H^{-1}g
\end{equation}
Since we need to guarantee the convergence of Newton method, we added step size ($\eta$):
\begin{equation}
b=a-\eta H^{-1}g
\end{equation}
In our paper, we want to solve for $\Delta p$. We substitutes the parameters Equation (30) as following:
\begin{equation}
p_{t+\Delta t}=p_t+\eta H^{-1}\sum\limits_{x,y}[\nabla I]^T[T(x,y)-I(W(x,y;p)]
\end{equation}
We compute the Hessian matrix in each pixel value by
\begin{equation}
H=
  \left[\begin{matrix}
    I_{xx} & I_{xy}\\
    I_{yx} & I_{yy}\\
  \end{matrix}\right]
\end{equation}
Since Lucas Kanade method assumes constant velocity between two nearby frames, we simplify the Hessian matrix by sum of square of each component:
\begin{equation}
 H=
  \left[\begin{matrix}
    sumsqr(I_{xx}) & sumsqr(I_{xy})\\
    sumsqr(I_{yx}) & sumsqr(I_{yy})\\
  \end{matrix}\right]
\end{equation}
Then we normalized Hessian matrix
\begin{equation}
 H_{norm}=\frac {H}{\sum_{i,j} H(i,j)}
\end{equation}
\newline
The Hessian matrix in Newton's method is directly computed. We compared the result with approximated Hessian matrix from Gauss-Newton method in Lucas Kanade method in experiment section.
\newline

\subsection{Inverse Compositional Algorithm for Lucas Kanede}
Traditional Lucas Kanade method in section 3 requires approximation of Hessian matrix for every iteration. \cite{6} proposed an inverse compositional algorithm by modifying the cost function as following:
\begin{equation}
\sum\limits_{x,y} [T(W(x,y;\Delta p))-I(W(x,y;p))]^2
\end{equation}
The first order term of Taylor series for Equation (33) is
\begin{equation}
\sum\limits_{x,y} [T(W(x,y;0))+\nabla T\frac{\partial W}{\partial p}\Delta p-I(W(x,y;p))]^2
\end{equation}
$W(x,y;0)$ is an identity warp
\begin{equation}
\sum\limits_{x,y} [T(x,y)+\nabla T\frac{\partial W}{\partial p}\Delta p-I(W(x,y;p))]^2
\end{equation}
The solution for least square problem in Equation (36) is
\begin{equation}
\Delta p=H^{-1}\sum\limits_{x,y}[\nabla T\frac{\partial W}{\partial p}]^T[I(W(x,y;p)-T(x,y)]
\end{equation}
\begin{equation}
H=\sum\limits_{x,y}[\nabla T\frac{\partial W}{\partial p}]^T[\nabla T\frac{\partial W}{\partial p}]
\end{equation}
From Equation (12)
\begin{equation}
\Delta p=H^{-1}\sum\limits_{x,y}[\nabla T]^T[I(W(x,y;p)-T(x,y)]
\end{equation}
\begin{equation}
H=\sum\limits_{x,y}[\nabla T]^T[\nabla T]
\end{equation}
Equation (40) shows that the Hessian matrix only depends on template image which will be initialized at the pre-computed stage. This means the computational cost for Hessian matrix will not significantly affect the optical flow estimation because it also need to be computed only once. Therefore, for second order optimization method, we experimented on Gauss-Newton method as the baseline from Lucas Kanade method, and Newton method for direct computation of Hessian matrix.
\newline

\section{Experiment}
Defining the gold standard for optical flow is a hard problem because human can not distinguish the movement of single pixel. We instead validated the performance of methods in section 2 on object tracking, an application of optical flow. Gold standard of tracking application can be manually observed by looking at the entire object and define its corner for each frame. We fixed the tracking box size for each object, and computed the error metric by Euclidean distance.
\subsection{Test Cases}
We generate 115 test cases (11,500 frames) to compare accuracy and speed of methods in section 2. Synthesis test cases were created by predefined shapes and convex hull from random points. Real world test cases were gathered from Youtube footage video.

\subsubsection{Synthetic Test Cases}
We first created black template with resolution 200x200 pixels, and added salt and pepper noise \cite{7} with density = 0.01. We generated 3 object size/object type : 15x15, 20x20, 25x25 pixels. The object types consist of:
\begin{enumerate}
	\item{Predefined shapes : circle, rectangle, triangle, hexagram (Figure 1)}
	\item{Convex hull of 5, 7, 9 random generated points (Figure 2)}
\end{enumerate}

\begin{figure}[!htbp]
	\centering
	\includegraphics[height=35 mm]{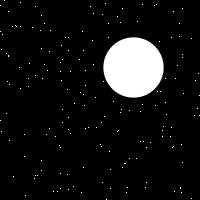}
	\includegraphics[height=35 mm]{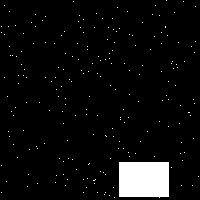}
	\includegraphics[height=35 mm]{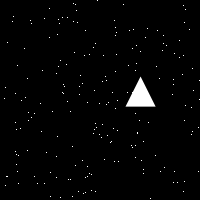}
	\includegraphics[height=35 mm]{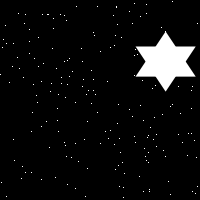}
	\caption{Synthetic test cases with predefined shapes}
\end{figure}

\begin{figure}[!htbp]
	\centering
	\includegraphics[height=35 mm]{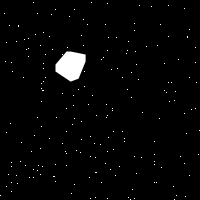}
	\includegraphics[height=35 mm]{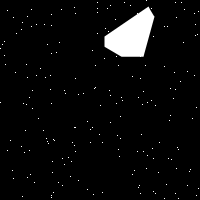}
	\includegraphics[height=35 mm]{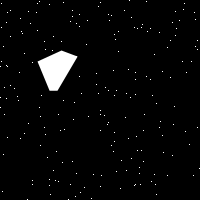}
	\includegraphics[height=35 mm]{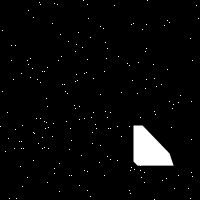}
	\caption{Synthetic test cases with convex hull}
\end{figure}

We solved convex hull by monotone chain \cite{8} and optimized the method by removing points within polygon generated by maximum and minimum points. 5 videos of 7 objects (with 3 size each) were created with 100 frames/video. In conclusion, we generated 105 videos with 10,500 frames for synthesis test cases. Tracking boxes were calculated from minimum and maximum coordinates of generated objects. We used these tracking boxes as gold standards, and we passed the first box as initial input for tracking algorithm.

\subsubsection{Real World Test Cases}
We used Bonn Benchmark on Tracking (BoBoT) \cite{9} dataset for real world test cases. We extracted 10 videos with 100 frames/video which consists of indoor objects and human (Figure 3). Each video has tracking obstacles, for instance, moving camera, background changes, rotation changes, scale changes, viewpoint changes. The BoBoT provides ground truth tracking boxes with dynamic size. We normalized tracking box by resizing them to $(min+max)/2$ and adjusted their position to the center of ground truth tracking boxes. 
\\

\begin{figure}[!htbp]
	\centering
	\includegraphics[height=27 mm]{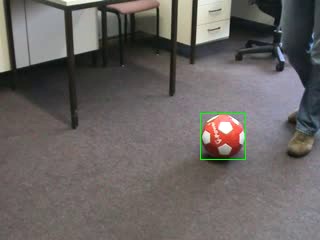}
	\includegraphics[height=27 mm]{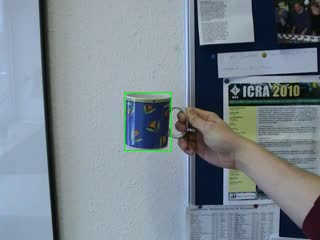}
	\includegraphics[height=27 mm]{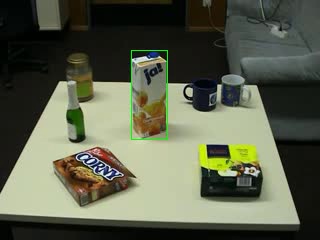}
	\includegraphics[height=27 mm]{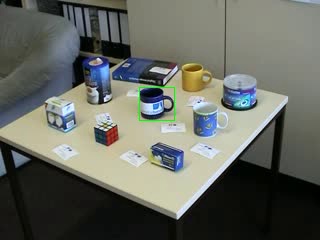}
	\caption{Real world test cases}
\end{figure}

\subsection{Results}
We provided experimental results for synthetic test cases (Table 1-3), and real world test cases (Table 4-6). The results were plotted in Figure 4-5. We computed average error for the tracking box with Euclidean distance$\leq$10 pixel, and counted the latter for fail frames. In synthetic test cases, Gauss-Newton method achieved the lowest average error, average time, and fail frames. The second best method, Gradient Descent with step size 0.02 has 0.054$\%$ higher error and 1.766 times slower.

\begin{table}[!htbp]
\centering
\caption{Average error (pixel) for synthetic test cases}
\label{my-label}
\begin{tabular}{|l|c|c|c|c|c|}
\hline
\multicolumn{1}{|c|}{\textbf{Method \textbackslash Step}} & \textbf{0.005} & \textbf{0.01} & \textbf{0.02} & \textbf{0.04} & \textbf{0.08} \\ \hline
Gauss-Newton & \multicolumn{5}{c|}{0.0614} \\ \hline
Gradient Descent & 0.1415 & 0.0618 & 0.0614 & 0.1254 & 1.4633 \\ \hline
Conjugate - Fletcher & 0.1056 & 0.0623 & 0.0693 & 0.0898 & 1.6338 \\ \hline
Conjugate - Polak & 0.1711 & 0.0625 & 0.1639 & 0.1967 & 2.6599 \\ \hline
Conjugate - Hestenes & 0.8177 & 0.3612 & 0.3575 & 0.8883 & 2.0041 \\ \hline
Conjugate - Dai Yun & 1.429 & 2.8531 & 0.583 & 0.6836 & 6.2065 \\ \hline
Newton's Method & 0.0623 & 0.1601 & 1.1828 & 3.0661 & 5.8747 \\ \hline
\end{tabular}
\end{table}

\begin{table}[!htbp]
\centering
\caption{Average time (s) for synthetic test cases}
\label{my-label}
\begin{tabular}{|l|c|c|c|c|c|}
\hline
\multicolumn{1}{|c|}{\textbf{Method \textbackslash Step size}} & \textbf{0.005} & \textbf{0.01} & \textbf{0.02} & \textbf{0.04} & \textbf{0.08} \\ \hline
Gauss-Newton & \multicolumn{5}{c|}{0.0077} \\ \hline
Gradient Descent & 0.0431 & 0.0259 & 0.0136 & 0.023 & 0.067 \\ \hline
Conjugate - Fletcher & 0.0384 & 0.0216 & 0.0115 & 0.0222 & 0.0526 \\ \hline
Conjugate - Polak & 0.0373 & 0.0218 & 0.0118 & 0.0153 & 0.0447 \\ \hline
Conjugate - Hestenes & 0.0332 & 0.0214 & 0.0119 & 0.0196 & 0.0233 \\ \hline
Conjugate - Dai Yun & 0.0241 & 0.0185 & 0.0192 & 0.02 & 0.0244 \\ \hline
Newton's Method & 0.0196 & 0.0186 & 0.0512 & 0.0583 & 0.0569 \\ \hline
\end{tabular}
\end{table}

\begin{table}[!htbp]
\centering
\caption{Fail frames (percentage) for synthetic test cases}
\label{my-label}
\begin{tabular}{|l|c|c|c|c|c|}
\hline
\multicolumn{1}{|c|}{\textbf{Method \textbackslash Step size}} & \textbf{0.005} & \textbf{0.01} & \textbf{0.02} & \textbf{0.04} & \textbf{0.08} \\ \hline
Gauss-Newton & \multicolumn{5}{c|}{0} \\ \hline
Gradient Descent & 9.5238 & 0 & 0 & 0 & 0 \\ \hline
Conjugate - Fletcher & 0.9524 & 0 & 1.9048 & 2.8571 & 41.905 \\ \hline
Conjugate - Polak & 8.5714 & 0 & 2.8571 & 18.095 & 69.524 \\ \hline
Conjugate - Hestenes & 40 & 26.667 & 23.81 & 36.19 & 94.286 \\ \hline
Conjugate - Dai Yun & 97.143 & 94.286 & 98.095 & 98.095 & 98.095 \\ \hline
Newton's Method & 0 & 0 & 0.9523 & 0.9523 & 8.5714 \\ \hline
\end{tabular}
\end{table}

In real world test cases, conjugate gradient descent with Dai Yun's $\beta$ achieved the lowest average error, however, it also has high failure percentage. We chose the second best average error, conjugate gradient descent with Fletcher Reevee's $\beta$, step size 0.08 which outperformed standard Gauss-Newton method by 56.977\% but also 1.551 times slower. In term of speed, Gauss-Newton method achieved fastest average time with at least 1.498 times faster than every other methods in our experiment.

\begin{table}[!htbp]
\centering
\caption{Average error (pixel) for real world test cases}
\label{my-label}
\begin{tabular}{|l|c|c|c|c|c|}
\hline
\multicolumn{1}{|c|}{\textbf{Method \textbackslash Step size}} & \textbf{0.005} & \textbf{0.01} & \textbf{0.02} & \textbf{0.04} & \textbf{0.08} \\ \hline
Gauss-Newton & \multicolumn{5}{c|}{2.765} \\ \hline
Gradient Descent & 2.4432 & 2.5667 & 2.233 & 2.2141 & 2.2398 \\ \hline
Conjugate - Fletcher & 2.3401 & 2.4317 & 2.2175 & 1.7637 & 1.7614 \\ \hline
Conjugate - Polak & 2.4466 & 2.5658 & 2.1814 & 2.2155 & 2.2435 \\ \hline
Conjugate - Hestenes & 2.5574 & 2.7499 & 2.2549 & 1.9914 & 2.2743 \\ \hline
Conjugate - Dai Yun & 3.0348 & 3.0517 & 2.5487 & 2.6038 & 0.6734 \\ \hline
Newton's Method & 2.8902 & 2.3305 & 2.2137 & 2.2596 & 3.2031 \\ \hline
\end{tabular}
\end{table}

\begin{table}[!htbp]
\centering
\caption{Average time (s) for real world test cases}
\label{my-label}
\begin{tabular}{|l|c|c|c|c|c|}
\hline
\multicolumn{1}{|c|}{\textbf{Method \textbackslash Step size}} & \textbf{0.005} & \textbf{0.01} & \textbf{0.02} & \textbf{0.04} & \textbf{0.08} \\ \hline
Gauss-Newton & \multicolumn{5}{c|}{0.0223} \\ \hline
Gradient Descent & 0.0741 & 0.0782 & 0.0593 & 0.0471 & 0.0334 \\ \hline
Conjugate - Fletcher & 0.0867 & 0.0793 & 0.0708 & 0.0482 & 0.0346 \\ \hline
Conjugate - Polak & 0.0948 & 0.081 & 0.0665 & 0.0523 & 0.0409 \\ \hline
Conjugate - Hestenes & 0.0849 & 0.0832 & 0.0715 & 0.0484 & 0.0398 \\ \hline
Conjugate - Dai Yun & 0.0683 & 0.0666 & 0.0637 & 0.0503 & 0.0312 \\ \hline
Newton's Method & 0.0666 & 0.0603 & 0.043 & 0.0378 & 0.0519 \\ \hline
\end{tabular}
\end{table}

\begin{table}[!htbp]
\centering
\caption{Fail frames (percentage) for real world test cases}
\label{my-label}
\begin{tabular}{|l|c|c|c|c|c|}
\hline
\multicolumn{1}{|c|}{\textbf{Method \textbackslash Step}} & \textbf{0.005} & \textbf{0.01} & \textbf{0.02} & \textbf{0.04} & \textbf{0.08} \\ \hline
Gauss-Newton & \multicolumn{5}{c|}{20} \\ \hline
Gradient Descent & 50 & 30 & 20 & 20 & 20 \\ \hline
Conjugate - Fletcher & 50 & 30 & 20 & 30 & 30 \\ \hline
Conjugate - Polak & 50 & 30 & 20 & 20 & 20 \\ \hline
Conjugate - Hestenes & 50 & 30 & 40 & 40 & 60 \\ \hline
Conjugate - Dai Yun & 50 & 50 & 50 & 50 & 70 \\ \hline
Newton's Method & 30 & 30 & 20 & 20 & 20 \\ \hline
\end{tabular}
\end{table}

Although Gauss-Newton outperformed every other method in synthetic test cases, the test cases did not contain tracking obstacle. Therefore, it contains bias from the model uniformity. We concluded conjugate gradient descent with Fletcher Reevee's $\beta$, step size 0.08 as an alternative method for higher accuracy. There are several methods with lower average error, however, optimized step size is needed for each dataset. This is the reason for Lucas-Kanade to choose Gauss-Newton as the optimization method.

\begin{figure}[!htbp]
	\centering
	\includegraphics[height=35 mm]{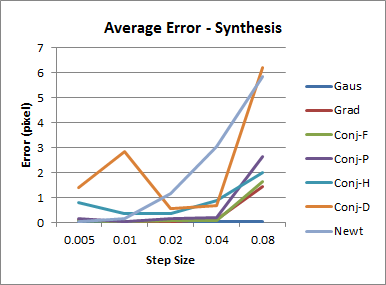} 
	\includegraphics[height=35 mm]{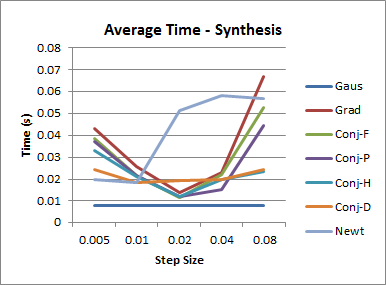} 
	\includegraphics[height=35 mm]{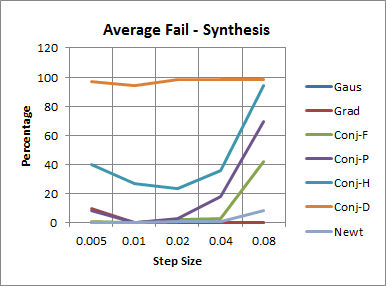}
\caption{Graph from synthetic test cases}
\end{figure}

\begin{figure}[!htbp]
	\centering
	\includegraphics[height=35 mm]{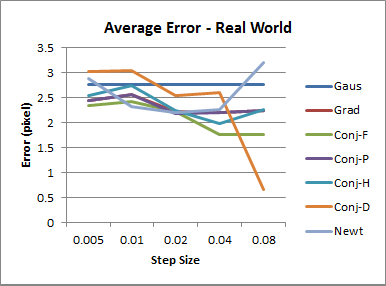} 
	\includegraphics[height=35 mm]{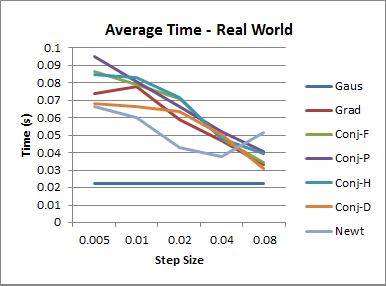} 
	\includegraphics[height=35 mm]{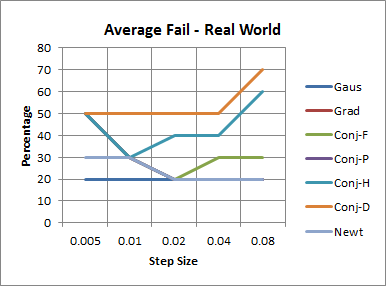}
\caption{Graph from real world test cases}
\end{figure}

\section{Conclusion}
This paper presents comparison of optimization methods in Lucas Kanade, a traditional optical flow estimation method. Comparison of four methods is done, including gradient descent, conjugate gradient descent, Newton's method, and Gauss-Newton method. Evaluation is achieved by calculation of Euclidean distance as the error metrics, tested on 10,500 synthetic frames and 1,000 real world footage frames. The experiment on synthetic test cases shows that Gauss-Newton method outperformed every other methods in both speed and accuracy. However, real world test cases image suggests conjugate gradient descent with Fletcher Reevee's $\beta$ has lowest error but also slower than Gauss-Newton method. In conclusion, training the right step size for conjugate gradient descent will minimize the Euclidean distance error, while Gauss-Newton method has highest speed and auto-calculated step size.

~\\
\subsection*{Contribution}
\begin{table}[!htbp]
\centering
\label{my-label}
\begin{tabular}{|l|l|l|l|l|l|l|l|l|l|l|l|l|l|}

\hline
\textbf{Section} & \multicolumn{1}{c|}{\textbf{1}}& \multicolumn{1}{c|}{\textbf{2}} & \multicolumn{1}{c|}{\textbf{2.1}} & \multicolumn{1}{c|}{\textbf{2.2}} & \multicolumn{1}{c|}{\textbf{2.3}} & \multicolumn{1}{c|}{\textbf{2.4}} & \multicolumn{1}{c|}{\textbf{2.5}} & \multicolumn{1}{c|}{\textbf{2.6}} &\multicolumn{1}{c|}{\textbf{3}}& \multicolumn{1}{c|}{\textbf{3.1.1}} & \multicolumn{1}{c|}{\textbf{3.1.2}} & \multicolumn{1}{c|}{\textbf{3.2}} & \multicolumn{1}{c|}{\textbf{4}} \\ \hline
\textbf{Noranart} &\multicolumn{1}{c|}{\ding{51}}& &\multicolumn{1}{c|}{\ding{51}} & & &\multicolumn{1}{c|}{\ding{51}} &\multicolumn{1}{c|}{\ding{51}}  & &\multicolumn{1}{c|}{\ding{51}}&\multicolumn{1}{c|}{\ding{51}} & &\multicolumn{1}{c|}{\ding{51}} &\multicolumn{1}{c|}{\ding{51}} \\ \hline
\textbf{Utkarsh} &\multicolumn{1}{c|}{\ding{51}}&\multicolumn{1}{c|}{\ding{51}} & &\multicolumn{1}{c|}{\ding{51}} &\multicolumn{1}{c|}{\ding{51}} & & &\multicolumn{1}{c|}{\ding{51}} && &\multicolumn{1}{c|}{\ding{51}} &\multicolumn{1}{c|}{\ding{51}} &\multicolumn{1}{c|}{\ding{51}} \\ \hline
\end{tabular}
\end{table}


\end{document}